\pdfoutput=1

\documentclass[11pt]{article}

\usepackage[final]{emnlp2021}

\usepackage{times}
\usepackage{latexsym}

\usepackage[T1]{fontenc}

\usepackage[utf8]{inputenc}

\usepackage{microtype}

\usepackage{times}
\usepackage{latexsym}
\usepackage{booktabs}
\usepackage{courier}  
\usepackage{helvet}  
\usepackage{tabularx}
\usepackage{url}
\usepackage{comment}
\usepackage{multirow}
\usepackage{amsmath} 
\usepackage{graphicx}
\usepackage{subfigure}
\usepackage[noend]{algpseudocode}
\usepackage{xspace}
\usepackage{threeparttable}
\usepackage{stmaryrd}
\usepackage{color}
\usepackage{stmaryrd}
\usepackage{xspace}
\usepackage{threeparttable}

\newcommand{\ignore}[1]{}
\usepackage{amsmath}
\usepackage{amssymb}
\usepackage{mathabx}
\usepackage{arydshln}

\usepackage[utf8]{inputenc}
\usepackage[T1]{fontenc}

\usepackage{xcolor}

\usepackage[normalem]{ulem}
\usepackage[ruled,vlined, linesnumbered]{algorithm2e}

\title{Knowledge Distillation with Noisy Labels for \\Natural Language Understanding}

\author{Shivendra Bhardwaj$^1$\thanks{\hspace{2mm}This work has been done while Shivendra Bhardwaj was at Huawei.} \hspace{3mm} Abbas Ghaddar$^1$ \hspace{3mm} Ahmad Rashid$^1$ \hspace{3mm} Khalil Bibi$^1$ \\ \textbf{Chengyang Li$^{1,2}$\thanks{\hspace{2mm}This work has been done while Chengyang Li was an intern at Huawei.} \hspace{3mm} Ali Ghodsi$^2$ \hspace{3mm} Philippe Langlais$^3$ \hspace{3mm} Mehdi Rezagholizadeh$^1$} \\
$^1$Huawei Noah’s Ark Lab\\
$^2$David R. Cheriton School of Computer Science, University of Waterloo\\
$^3$ RALI/DIRO, Universit\'e de Montr\'eal, Canada\\
\{abbas.ghaddar, ahmad.rashid,khalil.bibi, mehdi.rezagholizadeh\}@huawei.com\\ali.ghodsi@uwaterloo.ca, felipe@iro.umontreal.ca
}

\newcommand{\bert}{\textsc{Bert}}

\usepackage{ulem}

\newcommand{\mightmention}[1]{}
\newcommand{\problem}[1]{\textcolor{red}{$\star$}}
\newcommand{\answer}[1]{\textcolor{blue}{$\#$}}
\newcommand{\todoreview}[1]{\textcolor{green}{$@$}}

\begin{document}
 \maketitle
\begin{abstract}

Knowledge Distillation (KD) is extensively used to compress and deploy large pre-trained language models on edge devices for real-world applications. However, one neglected area of research is the impact of noisy (corrupted) labels on KD. We present, to the best of our knowledge, the first study on KD with noisy labels in Natural Language Understanding (NLU). We document the scope of the problem and present two methods to mitigate the impact of label noise. Experiments on the GLUE benchmark show that our methods are effective even under high noise levels. Nevertheless, our results indicate that more research is necessary to cope with label noise under the KD.

\end{abstract}

\section{Introduction}

Large-scale pre-trained language models~\cite{devlin2019bert,raffel2020exploring,brown2020language} have shown remarkable abilities to match and even surpass human performances on many Natural Languages Understanding (NLU) tasks~\cite{rajpurkar2018know,wang2018glue,wang2019superglue}. However, the deployment of these models in dynamic commercial environments come with challenges, including: large model size, and low training data quality. 

Knowledge Distillation~\cite{hinton2015distilling,turc2019well} is a compression technique of choice that has proven to be effective to fit a cumbersome NLU model on edge devices~\cite{sanh2019distilbert,jiao2020tinybert,sun2020mobilebert}. Meanwhile, numerous methods were developed to combat noisy (corrupted) labels, mainly for computer vision~\cite{frenay2013classification,jiang2018mentornet,thulasidasan2019combating,han2020survey} and more recently for NLU~\cite{ardehaly2018learning,jindal2019effective,garg2021towards,ghaddar2021context,ghaddar-etal-2021-end,jafari-etal-2021-annealing}.

Despite its success, KD has mostly been studied with the availability of massive amount of high quality labeled data. In practice, however, it is costly and impractical to produce such data~\cite{ghaddar2019contextualized}, and noisy labels are commonly encountered. In this paper, we consider the problem of KD when noisy labels are provided for training the main (teacher) and compressed (student) models. To our knowledge, this is the first time KD is studied under a noisy setting in NLU.

We conduct experiments on 7 tasks from the GLUE benchmark~\cite{wang2018glue} and observe a drastic drop of performance of distilled models when we increase the level of noise. In response, we propose 2 distillation training methods, namely Co-Distill and Label Refining, that are specifically designed to handle noise. Experiments show that our methods lead to improvements over fair baselines, and that it combination also performs the best. Yet, our analysis indicates that the problem is far from solved, and that there is much room for research. 

\section{Related Work}

The vanilla KD framework ~\cite{bucilua2006model,hinton2015distilling}  consists in training a small \textit{student} model to mimic the output of a large \textit{teacher} model. Recent years have seen a wide array of methods that leverage intermediate layer matching~\cite{ji2021show, wu2020skip,passban2020alp,wang2020minilm},  data augmentation~\cite{fu2020role,li2021select,jiao2020tinybert,kamalloo2021not}, or adversarial training~\cite{zaharia2021dialect,rashid2020towards,rashid2021matekd} in order to reduce the teacher-student performance gap. Instead, our proposed methods are designed to handle label noise during KD. Nevertheless, they can be easily fused with the aforementioned methods to further boost performance. 

Label noise (corruption) is a common problem in real-world datasets, and it has been well studied in the literature~\cite{frenay2013classification,li2017webvision,han2020survey}. Methods to combat noise build on the idea that samples with small training loss at early epochs are more likely to be clean~\cite{dehghani2018fidelity,wang2019symmetric}.

In co-teaching~\cite{han2018co}, two networks of different capacity teach each other to reject wrong labels. At each forward pass, each network keeps only small-loss samples and sends them to its peer network for updating the parameters. The main idea is that the error flow can be reduced, as networks of different learning abilities have different views on the data.

Self-distillation was proposed by \citet{dong2019distillation}, where the model is trained to mimic its own prediction from the previous training epoch. The goal is to prevent the model from memorizing wrong labels, as the model has less tendency to fit noise at early epochs. In addition, \citet{bagherinezhad2018label} showed improvements when distillation at early epochs is used to refine noisy labels.

Another line of works is the learning to weight approach~\cite{ren2018learning,li2019learning,zhang2020distilling,fan2020learning} that aims to learn per-sample loss weights in order to discount noisy samples. The proposed methods use an auxiliary meta-learner to re-weight training samples of the main model. However, all aforementioned works mainly focus on computer vision. Recently, \citet{garg2021towards} utilize a noise detection model to cluster, then score the training samples for text classification in an attempt to  guide the main model to focus on samples that are most likely to be correct.

\section{Methodology}

We first introduce our method, Co-Distill (CD), which  jointly trains the teacher and the student. Next, we incorporate Label Refinement (LR) which is motivated by the algorithms of \citet{jiang2018mentornet}, \citet{arazo2019unsupervised} and \citet{garg2021towards} for noise mitigation in regular (no KD) training framework.  

\subsection{Co-Distill (CD)}

\begin{figure}[htpb]
    \centering
    \includegraphics[width=0.48\textwidth]{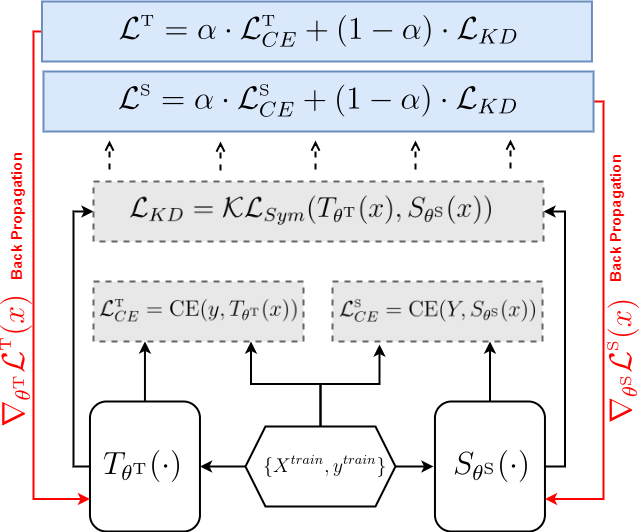}
    \caption{The Co-Distill architecture}
    \label{fig:ct}
\end{figure}

The key feature of our method is that the teacher and the student are trained together, but unlike traditional KD,  the teacher also learns from the student. Figure~\ref{fig:ct} showcases the complete architecture. 
We train the student model $S_{\theta^\text{\tiny{S}}}(\cdot)$ with the following loss function $\mathcal{L}^\text{\tiny{S}}$ :

 \begin{align}
    \label{eq:loss_ct}
    \mathcal{L^\text{\tiny{S}}} = \frac{1}{N}\sum_{i=1}^N  [&\alpha \cdot \mathcal{L}_{CE}(y_i, S_{\theta^\text{\tiny{S}}}(x_i)) \\\nonumber + & (1-\alpha)\cdot\mathcal{L}_{KD}(T_{\theta^\text{\tiny{T}}}(x_i), S_{\theta^\text{\tiny{S}}}(x_i))] \\\nonumber
    \end{align}

where $\theta^\text{\tiny{T}}$ and $\theta^\text{\tiny{S}}$ are the teacher and student parameters respectively, $\alpha$ is the KD weight parameter, $\mathcal{L}_\text{\tiny{CE}}$ is the Cross Entropy (CE) loss, $y_i$ is the label, $N$ is the total number of training samples and $\mathcal{L}_{KD}$ is the symmetric Kullback-Leibler (KL) divergence~\cite{kullback1997information} between the teacher and the student logits, i.e. we sum both the forward and reverse KL.

In addition to CE loss, the teacher "learns" from the student and is trained to minimize the $\mathcal{L}_{KD}$ loss. It is worth mentioning that we always train the teacher at the first epoch with an $\alpha$ value of 1. We do so to avoid propagating low confident information to the teacher at the beginning of the training. After the first epoch, the feedback of $\mathcal{L}_{KD}$ improves the overall performance of both teacher and student models.

\begin{table*}[!th]
    \centering
    \begin{tabular}{l l l l l l l lc}
    \toprule
         \textbf{Model}  & \textbf{CoLA} & \textbf{SST-2} & \textbf{MRPC} & \textbf{RTE} & \textbf{QNLI} & \textbf{QQP} & \textbf{MNLI} & \textbf{Avg.}  \\
         
         \midrule
         \multicolumn{9}{c}{\textit{0\%}}\\
         \midrule
         BERT-base &61.9 &93.1 &90.9 &68.6 &91.6 &91.6 &85.0 &83.2 \\
         \midrule
         w/o KD & 51.3  & 91.3 &87.5 &59.9 &89.2 &88.5 & 82.1&78.5 \\
         Vanilla & 56.4 &92.0 &90.0 &68.6 &90.3 &90.6 &85.0 &81.8 \\
         
         \midrule
         \multicolumn{9}{c}{\textit{25\%}}\\
         \midrule
         BERT-base & 46.7 &91.5 &75.7 &61.0 &87.9 &72.4 &81.8 &73.9  \\
         \hspace{2mm} with CD & 47.5 & 93.2 & 78.7 &62.5 &88.2 &71.3 &82.8 &74.9 \\
         \hspace{2mm} with CD+LR & 48.6 &92.8 &78.7 &60.3 &88.6 &74.0 &82.3 &75.0 \\
         \hdashline
         w/o KD &39.1 &90.4 &79.9 &61.0 &84.5 &67.3 &79.3 &71.6  \\
         Self-DSTL & 43.5 &90.5 &79.7 &60.6 &84.1 &69.3 &80.0 &72.5  \\ 
         Vanilla & 40.2 &90.9 &79.2 &61.7 &85.9 &72.9 &80.3 & 73.0 \\
         CD &45.1 &90.6 &79.2 &63.2 &85.5 &70.5 &80.7 & 73.5  \\ 
         CD+LR &\textbf{46.1} &\textbf{91.3} &\textbf{80.9} &\textbf{63.5} &\textbf{86.9} &\textbf{73.8} &\textbf{81.0} &\textbf{74.8} \\ 
         \midrule
         \multicolumn{9}{c}{\textit{50\%}}\\
         \midrule
         BERT-base & 17.7 &56.7 &68.4 &59.2 &64.7 &63.6 &76.5 &58.1  \\
         \hspace{2mm} with CD &16.0 &56.5 &70.8 &55.6 &62.7 &71.3 &76.8 &58.5 \\
         \hspace{2mm} with CD+LR & 17.2& 61.7& 71.8& 56.3& 62.7 &71.3 & 77.0 & 59.7 \\
         \hdashline
         w/o KD &8.3 &55.0 &66.6 &57.0 &56.5 &67.3 &72.2 &54.7 \\
         Self-DSTL &8.8 &57.6 &68.1 &58.4 &57.3 &69.3 &73.2 &56.1  \\
         Vanilla &11.9 &60.0 &68.6 &\textbf{58.5} &60.3 &66.1 &75.0 &57.2\\

         CD &13.6 &60.3 &69.1 &57.4 &60.0 &70.5 &75.1 &  58.0\\ 
         CD+LR &\textbf{17.7} &\textbf{64.1} &\textbf{71.1} &57.4 &\textbf{60.9} &\textbf{73.8} &\textbf{76.6} & \textbf{60.2}  \\ 
        
        \bottomrule
    \end{tabular}
    
    \caption{Performances on GLUE dev sets of models trained on 0\%, 25\%, and 50\% of noisy labels. Dash lines separate teacher (up) and student models.}
    
    \label{tab:glue_main}
\end{table*}

\subsection{CD plus Label Refinement (CD+LR)}

We further enhance CD by refining the training labels based on loss values at early epochs. In LR, an auxiliary classifier is trained to flag  noisy samples, which in turn are re-labeled by the main model. In prior work on noisy labels~\cite{arpit2017closer,dehghani2018fidelity,wang2019symmetric} it has been observed that small training losses at early epochs are more likely to indicate that a sample is clean. 

Instead, we assume that we have access to a small subset of validation data where noisy and clean samples are known a priori (see Section~\ref{sec:Dataset and Evaluation}). We train both teacher and student with Co-Distill for 2 epochs,\footnote{Empirically, we found that  it works well on most of the tasks we experimented on.} and then calculate $\mathcal{L}^{T}_{CE}$ and $\mathcal{L}^{S}_{CE}$ for each sample in the validation set. 

We use these values as features for a discriminator model $D(.)$ trained to predict whether a sample is noisy. Once it is trained, $D(.)$ is used to flag noisy training samples, so that the teacher re-labels them. Finally, we resume the co-distillation for the remaining epochs while calculating the CE loss using the new labels.

\section{Experiments}

\subsection{Dataset and Evaluation}
\label{sec:Dataset and Evaluation}
We experiment on 7 tasks from the GLUE benchmark~\cite{wang2018glue}: 2 single-sentence (CoLA and SST-2) and 5 sentence-pair (MRPC, RTE, QQP, QNLI, and MNLI) classification tasks. Following prior work, we report Matthews correlation on CoLA and accuracy for the other tasks. Since GLUE test sets are hidden and the number of submissions to leaderboard is limited, we held-out 10\% of the training set for validation and used the rest for training. We used this validation set to train the discriminator as well as for hyper-parameter tuning, while official GLUE dev sets are used to evaluate the models.

We test our methods on training sets with 25\% and 50\% noisy labels\footnote{We do not evaluate beyond 50\% of noise because many GLUE tasks are binary classification.}. We introduce the same level of noise for the validation sets.    
Following prior works~\cite{jiang2018mentornet,dong2019distillation,garg2021towards}, we inject artificial noise by randomly changing the original labels of the training samples.

\subsection{Baselines}

We compare our noise mitigation methods with 3 popular baselines: 
\vspace{-1mm}
\begin{itemize}
    \item \textbf{w/o KD} In this setting, only the CE loss is used. This baseline is used as a witness.
    
    \item \textbf{Vanilla-KD} Here, we select the best performing $\alpha$ value for each task.
    
    \item \textbf{Self-DSTL} In Self-Distillation~\cite{dong2019distillation}, the student is first trained for few epochs on hard labels only, and the best checkpoint is used to generate logits on the training data. For the rest of the epochs the student is trained on both hard and its own soft labels. 

\end{itemize}

\subsection{Implementation}
We use as our teacher the 12-layer \bert{-base-uncased} model~\cite{devlin2019bert}, and the pre-trained 6-layer distillBERT~\cite{sanh2019distilbert} to initialize all student models. We use \texttt{scikit-learn}~\cite{pedregosa2011scikit} to train a Random Forest discriminator~\cite{breiman2001random} as our auxiliary classifier. For all models, we perform hyper-parameter tuning and best model selection based on early stopping on noisy validation sets. We report average results over 3 random seeds.

\subsection{Results}

Table~\ref{tab:glue_main} shows performances on GLUE dev sets of 3 teachers and 5 student models trained on clean (0\%), 25\% and 50\% of noisy training sets. As expected, the performance of all models drops drastically with noise. For instance, the teacher and vanilla student average performances drop by 25.1\% and 24.7\% respectively when we train with 50\% of noisy labels. Among all baselines, training the student solely on hard labels (w/o KD) performs the worst under all levels of noise. 

Performing distillation with the student logits itself (Self-DSTL) slightly improves the performances by 0.5\% and 1.5\% on 25\% and 50\% noise level respectively. However, using teacher logits (Vanilla) for distillation always performs better than using that of the student by 1\% on average. This indicates that the teacher knowledge remains crucial even under a noisy label setting.     

Overall, our method CD leads to an average gain of 0.5\% and 0.8\% on top of the Vanilla baseline at 25\% and 50\% noise level respectively. Moreover, enhancing the CD methods with label refinement (CD+LR) significantly boosts these scores by 1.3\% and 2.2\% respectively. CD+LR consistently outperforms Vanilla KD across all tasks and noise levels, except at 50\% noise for MRPC. It is worth noting that our methods are more effective under extreme noise level, since the gap with Vanilla KD gets larger at 50\% noise level (the max for binary classification).   

On the teacher side, we observe that the teachers obtained with our methods outperform the other teachers. The CD+LR teacher is better than its naive counterpart by 1.1\% and 1.6\% on 25\% and 50\% noise level respectively. This observation is inline with \newcite{han2018co} who find that in Co-teaching, the two networks communicating with each other get improved. More interestingly, results show that CD+LR students outperform significantly ($>$2\%) the naive and slightly (0.2\%) their respective teachers. This is mainly due to the tendency of over-parameterized neural networks (teachers) to fit noisy labels~\cite{han2018co,jiang2018mentornet}, compared to smaller models (students in our cases). This suggests that in a high noise setting, training a robust teacher is important as much as training the student.

\begin{figure}[!th]
    \centering
    \includegraphics[width=0.48\textwidth]{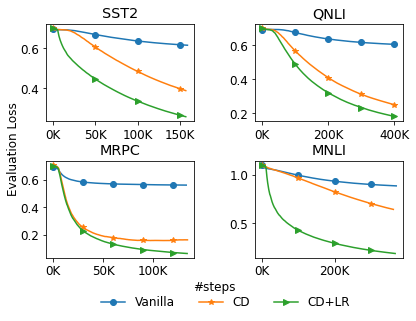}
    \caption{Validation (on GLUE dev sets) curve for the 3 student models trained on 50\% of noisy labels.}
    \label{fig:eval_loss}
\end{figure}

We plot the losses on dev sets at early steps to better understand how our methods combat noisy labels. Figure~\ref{fig:eval_loss} shows dev loss values on 4 GLUE tasks\footnote{Similar figures are observed on the remaining 3 tasks.} for Vanilla KD, CD, and CD+LR methods. First, we observe that the loss curve of Vanilla KD flattens at early stages. We investigated the training loss and noticed that it rather decreases, mainly due over-fitting the noise labels. 

Co-Distillation (CD) shows better signs of mitigating noise, as the loss  decreases slowly on MNLI and sharply on QNLI and MRPC. Adding LR leads to a sharp drop, followed by a steady decrease of loss values. The drop happens immediately after refining the training set labels, which seems crucial for large  datasets like MNLI and SST-2.  

\section{Conclusion}
We present the first study on Knowledge Distillation when learning from noisy labels in NLU, and show that the problem is extremely challenging. Future work involves conducting a comparative study on the robustness of state-of-the-art KD techniques against noisy labels, and merging them within our methods. We hope that our study will encourage future research on KD in the noisy label setting, a genuine setting in real world applications.

\section*{Acknowledgments}
We thank Mindspore\footnote{\url{https://www.mindspore.cn/}} for the partial support of this work, which is a new deep learning computing framework.

\bibliography{custom}

\begin{thebibliography}{48}
\expandafter\ifx\csname natexlab\endcsname\relax\def\natexlab#1{#1}\fi

\bibitem[{Arazo et~al.(2019)Arazo, Ortego, Albert, O’Connor, and
  McGuinness}]{arazo2019unsupervised}
Eric Arazo, Diego Ortego, Paul Albert, Noel O’Connor, and Kevin McGuinness.
  2019.
\newblock Unsupervised label noise modeling and loss correction.
\newblock In \emph{International Conference on Machine Learning}, pages
  312--321. PMLR.

\bibitem[{Ardehaly and Culotta(2018)}]{ardehaly2018learning}
Ehsan~Mohammady Ardehaly and Aron Culotta. 2018.
\newblock Learning from noisy label proportions for classifying online social
  data.
\newblock \emph{Social Network Analysis and Mining}, 8(1):1--18.

\bibitem[{Arpit et~al.(2017)Arpit, Jastrzębski, Ballas, Krueger, Bengio,
  Kanwal, Maharaj, Fischer, Courville, Bengio, and
  Lacoste-Julien}]{arpit2017closer}
Devansh Arpit, Stanisław Jastrzębski, Nicolas Ballas, David Krueger, Emmanuel
  Bengio, Maxinder~S. Kanwal, Tegan Maharaj, Asja Fischer, Aaron Courville,
  Yoshua Bengio, and Simon Lacoste-Julien. 2017.
\newblock \href {http://arxiv.org/abs/1706.05394} {A closer look at
  memorization in deep networks}.

\bibitem[{Bagherinezhad et~al.(2018)Bagherinezhad, Horton, Rastegari, and
  Farhadi}]{bagherinezhad2018label}
Hessam Bagherinezhad, Maxwell Horton, Mohammad Rastegari, and Ali Farhadi.
  2018.
\newblock Label refinery: Improving imagenet classification through label
  progression.
\newblock \emph{arXiv preprint arXiv:1805.02641}.

\bibitem[{Breiman(2001)}]{breiman2001random}
Leo Breiman. 2001.
\newblock Random forests.
\newblock \emph{Machine learning}, 45(1):5--32.

\bibitem[{Brown et~al.(2020)Brown, Mann, Ryder, Subbiah, Kaplan, Dhariwal,
  Neelakantan, Shyam, Sastry, Askell et~al.}]{brown2020language}
Tom~B Brown, Benjamin Mann, Nick Ryder, Melanie Subbiah, Jared Kaplan, Prafulla
  Dhariwal, Arvind Neelakantan, Pranav Shyam, Girish Sastry, Amanda Askell,
  et~al. 2020.
\newblock Language models are few-shot learners.
\newblock \emph{arXiv preprint arXiv:2005.14165}.

\bibitem[{Buciluǎ et~al.(2006)Buciluǎ, Caruana, and
  Niculescu-Mizil}]{bucilua2006model}
Cristian Buciluǎ, Rich Caruana, and Alexandru Niculescu-Mizil. 2006.
\newblock Model compression.
\newblock In \emph{Proceedings of the 12th ACM SIGKDD international conference
  on Knowledge discovery and data mining}, pages 535--541.

\bibitem[{Dehghani et~al.(2018)Dehghani, Mehrjou, Gouws, Kamps, and
  Sch{\"o}lkopf}]{dehghani2018fidelity}
Mostafa Dehghani, Arash Mehrjou, Stephan Gouws, Jaap Kamps, and Bernhard
  Sch{\"o}lkopf. 2018.
\newblock Fidelity-weighted learning.
\newblock In \emph{International Conference on Learning Representations}.

\bibitem[{Devlin et~al.(2019)Devlin, Chang, Lee, and
  Toutanova}]{devlin2019bert}
Jacob Devlin, Ming-Wei Chang, Kenton Lee, and Kristina Toutanova. 2019.
\newblock \href {http://arxiv.org/abs/1810.04805} {Bert: Pre-training of deep
  bidirectional transformers for language understanding}.

\bibitem[{Dong et~al.(2019)Dong, Hou, Lu, and Zhang}]{dong2019distillation}
Bin Dong, Jikai Hou, Yiping Lu, and Zhihua Zhang. 2019.
\newblock Distillation = early stopping? harvesting dark knowledge utilizing
  anisotropic information retrieval for overparameterized neural network.
\newblock \emph{stat}, 1050:2.

\bibitem[{Fan et~al.(2020)Fan, Xia, Wu, Xie, Liu, Bian, Qin, Li, and
  Liu}]{fan2020learning}
Yang Fan, Yingce Xia, Lijun Wu, Shufang Xie, Weiqing Liu, Jiang Bian, Tao Qin,
  Xiang-Yang Li, and Tie-Yan Liu. 2020.
\newblock Learning to teach with deep interactions.
\newblock \emph{arXiv preprint arXiv:2007.04649}.

\bibitem[{Fr{\'e}nay and Verleysen(2013)}]{frenay2013classification}
Beno{\^\i}t Fr{\'e}nay and Michel Verleysen. 2013.
\newblock Classification in the presence of label noise: a survey.
\newblock \emph{IEEE transactions on neural networks and learning systems},
  25(5):845--869.

\bibitem[{Fu et~al.(2020)Fu, Geng, Duan, Zhuang, Yuan, Trischler, Lin, Pal, and
  Dong}]{fu2020role}
Jie Fu, Xue Geng, Zhijian Duan, Bohan Zhuang, Xingdi Yuan, Adam Trischler, Jie
  Lin, Chris Pal, and Hao Dong. 2020.
\newblock Role-wise data augmentation for knowledge distillation.
\newblock \emph{arXiv preprint arXiv:2004.08861}.

\bibitem[{Garg et~al.(2021)Garg, Ramakrishnan, and Thumbe}]{garg2021towards}
Siddhant Garg, Goutham Ramakrishnan, and Varun Thumbe. 2021.
\newblock Towards robustness to label noise in text classification via noise
  modeling.
\newblock \emph{arXiv preprint arXiv:2101.11214}.

\bibitem[{Ghaddar and Langlais(2019)}]{ghaddar2019contextualized}
Abbas Ghaddar and Philippe Langlais. 2019.
\newblock Contextualized word representations from distant supervision with and
  for ner.
\newblock In \emph{Proceedings of the 5th Workshop on Noisy User-generated Text
  (W-NUT 2019)}, pages 101--108.

\bibitem[{Ghaddar et~al.(2021{\natexlab{a}})Ghaddar, Langlais, Rashid, and
  Rezagholizadeh}]{ghaddar2021context}
Abbas Ghaddar, Philippe Langlais, Ahmad Rashid, and Mehdi Rezagholizadeh.
  2021{\natexlab{a}}.
\newblock \href {https://transacl.org/ojs/index.php/tacl/article/view/2669}
  {Context-aware adversarial training for name regularity bias in named entity
  recognition}.
\newblock \emph{Trans. Assoc. Comput. Linguistics}, 9:586--604.

\bibitem[{Ghaddar et~al.(2021{\natexlab{b}})Ghaddar, Langlais, Rezagholizadeh,
  and Rashid}]{ghaddar-etal-2021-end}
Abbas Ghaddar, Philippe Langlais, Mehdi Rezagholizadeh, and Ahmad Rashid.
  2021{\natexlab{b}}.
\newblock \href {https://doi.org/10.18653/v1/2021.findings-acl.168} {End-to-end
  self-debiasing framework for robust {NLU} training}.
\newblock In \emph{Findings of the Association for Computational Linguistics:
  {ACL/IJCNLP} 2021, Online Event, August 1-6, 2021}, volume {ACL/IJCNLP} 2021
  of \emph{Findings of {ACL}}, pages 1923--1929. Association for Computational
  Linguistics.

\bibitem[{Han et~al.(2018)Han, Yao, Yu, Niu, Xu, Hu, Tsang, and
  Sugiyama}]{han2018co}
B~Han, Q~Yao, X~Yu, G~Niu, M~Xu, W~Hu, IW~Tsang, and M~Sugiyama. 2018.
\newblock Co-teaching: Robust training of deep neural networks with extremely
  noisy labels.
\newblock In \emph{32nd Conference on Neural Information Processing Systems
  (NIPS)}. NEURAL INFORMATION PROCESSING SYSTEMS (NIPS).

\bibitem[{Han et~al.(2020)Han, Yao, Liu, Niu, Tsang, Kwok, and
  Sugiyama}]{han2020survey}
Bo~Han, Quanming Yao, Tongliang Liu, Gang Niu, Ivor~W Tsang, James~T Kwok, and
  Masashi Sugiyama. 2020.
\newblock A survey of label-noise representation learning: Past, present and
  future.
\newblock \emph{arXiv preprint arXiv:2011.04406}.

\bibitem[{Hinton et~al.(2015)Hinton, Vinyals, and Dean}]{hinton2015distilling}
Geoffrey Hinton, Oriol Vinyals, and Jeffrey Dean. 2015.
\newblock \href {http://arxiv.org/abs/1503.02531} {Distilling the knowledge in
  a neural network}.
\newblock In \emph{NIPS Deep Learning and Representation Learning Workshop}.

\bibitem[{Jafari et~al.(2021)Jafari, Rezagholizadeh, Sharma, and
  Ghodsi}]{jafari-etal-2021-annealing}
Aref Jafari, Mehdi Rezagholizadeh, Pranav Sharma, and Ali Ghodsi. 2021.
\newblock \href {https://aclanthology.org/2021.eacl-main.212} {Annealing
  knowledge distillation}.
\newblock In \emph{Proceedings of the 16th Conference of the European Chapter
  of the Association for Computational Linguistics: Main Volume}, pages
  2493--2504, Online. Association for Computational Linguistics.

\bibitem[{Ji et~al.(2021)Ji, Heo, and Park}]{ji2021show}
Mingi Ji, Byeongho Heo, and Sungrae Park. 2021.
\newblock Show, attend and distill: Knowledge distillation via attention-based
  feature matching.
\newblock In \emph{Proceedings of the AAAI Conference on Artificial
  Intelligence}.

\bibitem[{Jiang et~al.(2018)Jiang, Zhou, Leung, Li, and
  Fei-Fei}]{jiang2018mentornet}
Lu~Jiang, Zhengyuan Zhou, Thomas Leung, Li-Jia Li, and Li~Fei-Fei. 2018.
\newblock Mentornet: Learning data-driven curriculum for very deep neural
  networks on corrupted labels.
\newblock In \emph{International Conference on Machine Learning}, pages
  2304--2313. PMLR.

\bibitem[{Jiao et~al.(2020)Jiao, Yin, Shang, Jiang, Chen, Li, Wang, and
  Liu}]{jiao2020tinybert}
Xiaoqi Jiao, Yichun Yin, Lifeng Shang, Xin Jiang, Xiao Chen, Linlin Li, Fang
  Wang, and Qun Liu. 2020.
\newblock Tinybert: Distilling bert for natural language understanding.
\newblock In \emph{Proceedings of the 2020 Conference on Empirical Methods in
  Natural Language Processing: Findings}, pages 4163--4174.

\bibitem[{Jindal et~al.(2019)Jindal, Pressel, Lester, and
  Nokleby}]{jindal2019effective}
Ishan Jindal, Daniel Pressel, Brian Lester, and Matthew Nokleby. 2019.
\newblock An effective label noise model for dnn text classification.
\newblock In \emph{Proceedings of the 2019 Conference of the North American
  Chapter of the Association for Computational Linguistics: Human Language
  Technologies, Volume 1 (Long and Short Papers)}, pages 3246--3256.

\bibitem[{Kamalloo et~al.(2021)Kamalloo, Rezagholizadeh, Passban, and
  Ghodsi}]{kamalloo2021not}
Ehsan Kamalloo, Mehdi Rezagholizadeh, Peyman Passban, and Ali Ghodsi. 2021.
\newblock \href {https://doi.org/10.18653/v1/2021.findings-acl.309} {Not far
  away, not so close: Sample efficient nearest neighbour data augmentation via
  minimax}.
\newblock In \emph{Findings of the Association for Computational Linguistics:
  {ACL/IJCNLP} 2021, Online Event, August 1-6, 2021}, volume {ACL/IJCNLP} 2021
  of \emph{Findings of {ACL}}, pages 3522--3533. Association for Computational
  Linguistics.

\bibitem[{Kullback(1997)}]{kullback1997information}
Solomon Kullback. 1997.
\newblock \emph{Information theory and statistics}.
\newblock Courier Corporation.

\bibitem[{Li et~al.(2019)Li, Wong, Zhao, and Kankanhalli}]{li2019learning}
Junnan Li, Yongkang Wong, Qi~Zhao, and Mohan~S Kankanhalli. 2019.
\newblock Learning to learn from noisy labeled data.
\newblock In \emph{Proceedings of the IEEE/CVF Conference on Computer Vision
  and Pattern Recognition}, pages 5051--5059.

\bibitem[{Li et~al.(2021)Li, Rashid, Jafari, Sharma, Ghodsi, and
  Rezagholizadeh}]{li2021select}
Tianda Li, Ahmad Rashid, Aref Jafari, Pranav Sharma, Ali Ghodsi, and Mehdi
  Rezagholizadeh. 2021.
\newblock How to select one among all? an extensive empirical study towards the
  robustness of knowledge distillation in natural language understanding.
\newblock \emph{arXiv preprint arXiv:2109.05696}.

\bibitem[{Li et~al.(2017)Li, Wang, Li, Agustsson, and
  Van~Gool}]{li2017webvision}
Wen Li, Limin Wang, Wei Li, Eirikur Agustsson, and Luc Van~Gool. 2017.
\newblock Webvision database: Visual learning and understanding from web data.
\newblock \emph{arXiv preprint arXiv:1708.02862}.

\bibitem[{Passban et~al.(2021)Passban, Wu, Rezagholizadeh, and
  Liu}]{passban2020alp}
Peyman Passban, Yimeng Wu, Mehdi Rezagholizadeh, and Qun Liu. 2021.
\newblock \href {https://ojs.aaai.org/index.php/AAAI/article/view/17610}
  {{ALP-KD:} attention-based layer projection for knowledge distillation}.
\newblock In \emph{Thirty-Fifth {AAAI} Conference on Artificial Intelligence,
  {AAAI} 2021, Thirty-Third Conference on Innovative Applications of Artificial
  Intelligence, {IAAI} 2021, The Eleventh Symposium on Educational Advances in
  Artificial Intelligence, {EAAI} 2021, Virtual Event, February 2-9, 2021},
  pages 13657--13665. {AAAI} Press.

\bibitem[{Pedregosa et~al.(2011)Pedregosa, Varoquaux, Gramfort, Michel,
  Thirion, Grisel, Blondel, Prettenhofer, Weiss, Dubourg
  et~al.}]{pedregosa2011scikit}
Fabian Pedregosa, Ga{\"e}l Varoquaux, Alexandre Gramfort, Vincent Michel,
  Bertrand Thirion, Olivier Grisel, Mathieu Blondel, Peter Prettenhofer, Ron
  Weiss, Vincent Dubourg, et~al. 2011.
\newblock Scikit-learn: Machine learning in python.
\newblock \emph{the Journal of machine Learning research}, 12:2825--2830.

\bibitem[{Raffel et~al.(2020)Raffel, Shazeer, Roberts, Lee, Narang, Matena,
  Zhou, Li, and Liu}]{raffel2020exploring}
Colin Raffel, Noam Shazeer, Adam Roberts, Katherine Lee, Sharan Narang, Michael
  Matena, Yanqi Zhou, Wei Li, and Peter~J Liu. 2020.
\newblock Exploring the limits of transfer learning with a unified text-to-text
  transformer.
\newblock \emph{Journal of Machine Learning Research}, 21:1--67.

\bibitem[{Rajpurkar et~al.(2018)Rajpurkar, Jia, and Liang}]{rajpurkar2018know}
Pranav Rajpurkar, Robin Jia, and Percy Liang. 2018.
\newblock Know what you don’t know: Unanswerable questions for squad.
\newblock In \emph{Proceedings of the 56th Annual Meeting of the Association
  for Computational Linguistics (Volume 2: Short Papers)}, pages 784--789.

\bibitem[{Rashid et~al.(2020)Rashid, Lioutas, Ghaddar, and
  Rezagholizadeh}]{rashid2020towards}
Ahmad Rashid, Vasileios Lioutas, Abbas Ghaddar, and Mehdi Rezagholizadeh. 2020.
\newblock \href {http://arxiv.org/abs/2012.15495} {Towards zero-shot knowledge
  distillation for natural language processing}.
\newblock \emph{CoRR}, abs/2012.15495.

\bibitem[{Rashid et~al.(2021)Rashid, Lioutas, and
  Rezagholizadeh}]{rashid2021matekd}
Ahmad Rashid, Vasileios Lioutas, and Mehdi Rezagholizadeh. 2021.
\newblock \href {https://doi.org/10.18653/v1/2021.acl-long.86} {{MATE}-{KD}:
  Masked adversarial {TE}xt, a companion to knowledge distillation}.
\newblock In \emph{Proceedings of the 59th Annual Meeting of the Association
  for Computational Linguistics and the 11th International Joint Conference on
  Natural Language Processing (Volume 1: Long Papers)}, pages 1062--1071,
  Online. Association for Computational Linguistics.

\bibitem[{Ren et~al.(2018)Ren, Zeng, Yang, and Urtasun}]{ren2018learning}
Mengye Ren, Wenyuan Zeng, Bin Yang, and Raquel Urtasun. 2018.
\newblock Learning to reweight examples for robust deep learning.
\newblock In \emph{International Conference on Machine Learning}, pages
  4334--4343. PMLR.

\bibitem[{Sanh et~al.(2019)Sanh, Debut, Chaumond, and
  Wolf}]{sanh2019distilbert}
Victor Sanh, Lysandre Debut, Julien Chaumond, and Thomas Wolf. 2019.
\newblock Distilbert, a distilled version of bert: smaller, faster, cheaper and
  lighter.
\newblock \emph{arXiv preprint arXiv:1910.01108}.

\bibitem[{Sun et~al.(2020)Sun, Yu, Song, Liu, Yang, and
  Zhou}]{sun2020mobilebert}
Zhiqing Sun, Hongkun Yu, Xiaodan Song, Renjie Liu, Yiming Yang, and Denny Zhou.
  2020.
\newblock Mobilebert: a compact task-agnostic bert for resource-limited
  devices.
\newblock In \emph{Proceedings of the 58th Annual Meeting of the Association
  for Computational Linguistics}, pages 2158--2170.

\bibitem[{Thulasidasan et~al.(2019)Thulasidasan, Bhattacharya, Bilmes,
  Chennupati, and Mohd-Yusof}]{thulasidasan2019combating}
Sunil Thulasidasan, Tanmoy Bhattacharya, Jeff Bilmes, Gopinath Chennupati, and
  Jamal Mohd-Yusof. 2019.
\newblock Combating label noise in deep learning using abstention.
\newblock \emph{arXiv preprint arXiv:1905.10964}.

\bibitem[{Turc et~al.(2019)Turc, Chang, Lee, and Toutanova}]{turc2019well}
Iulia Turc, Ming-Wei Chang, Kenton Lee, and Kristina Toutanova. 2019.
\newblock Well-read students learn better: On the importance of pre-training
  compact models.
\newblock \emph{arXiv preprint arXiv:1908.08962}.

\bibitem[{Wang et~al.(2019{\natexlab{a}})Wang, Pruksachatkun, Nangia, Singh,
  Michael, Hill, Levy, and Bowman}]{wang2019superglue}
Alex Wang, Yada Pruksachatkun, Nikita Nangia, Amanpreet Singh, Julian Michael,
  Felix Hill, Omer Levy, and Samuel~R Bowman. 2019{\natexlab{a}}.
\newblock Superglue: A stickier benchmark for general-purpose language
  understanding systems.
\newblock \emph{Advances in Neural Information Processing Systems}, 32.

\bibitem[{Wang et~al.(2018)Wang, Singh, Michael, Hill, Levy, and
  Bowman}]{wang2018glue}
Alex Wang, Amanpreet Singh, Julian Michael, Felix Hill, Omer Levy, and Samuel
  Bowman. 2018.
\newblock {GLUE: A Multi-Task Benchmark and Analysis Platform for Natural
  Language Understanding}.
\newblock In \emph{Proceedings of the 2018 EMNLP Workshop BlackboxNLP:
  Analyzing and Interpreting Neural Networks for NLP}, pages 353--355.

\bibitem[{Wang et~al.(2020)Wang, Wei, Dong, Bao, Yang, and
  Zhou}]{wang2020minilm}
Wenhui Wang, Furu Wei, Li~Dong, Hangbo Bao, Nan Yang, and Ming Zhou. 2020.
\newblock Minilm: Deep self-attention distillation for task-agnostic
  compression of pre-trained transformers.
\newblock \emph{arXiv preprint arXiv:2002.10957}.

\bibitem[{Wang et~al.(2019{\natexlab{b}})Wang, Ma, Chen, Luo, Yi, and
  Bailey}]{wang2019symmetric}
Yisen Wang, Xingjun Ma, Zaiyi Chen, Yuan Luo, Jinfeng Yi, and James Bailey.
  2019{\natexlab{b}}.
\newblock Symmetric cross entropy for robust learning with noisy labels.
\newblock In \emph{Proceedings of the IEEE/CVF International Conference on
  Computer Vision}, pages 322--330.

\bibitem[{Wu et~al.(2020)Wu, Passban, Rezagholizadeh, and Liu}]{wu2020skip}
Yimeng Wu, Peyman Passban, Mehdi Rezagholizadeh, and Qun Liu. 2020.
\newblock \href {https://doi.org/10.18653/v1/2020.emnlp-main.74} {Why skip if
  you can combine: {A} simple knowledge distillation technique for intermediate
  layers}.
\newblock In \emph{Proceedings of the 2020 Conference on Empirical Methods in
  Natural Language Processing, {EMNLP} 2020, Online, November 16-20, 2020},
  pages 1016--1021. Association for Computational Linguistics.

\bibitem[{Zaharia et~al.(2021)Zaharia, Avram, Cercel, and
  Rebedea}]{zaharia2021dialect}
George{-}Eduard Zaharia, Andrei{-}Marius Avram, Dumitru{-}Clementin Cercel, and
  Traian Rebedea. 2021.
\newblock \href {https://aclanthology.org/2021.vardial-1.13/} {Dialect
  identification through adversarial learning and knowledge distillation on
  romanian {BERT}}.
\newblock In \emph{Proceedings of the Eighth Workshop on {NLP} for Similar
  Languages, Varieties and Dialects, VarDial@EACL 2021, Kiyv, Ukraine, April
  20, 2021}, pages 113--119. Association for Computational Linguistics.

\bibitem[{Zhang et~al.(2020)Zhang, Zhang, Arik, Lee, and
  Pfister}]{zhang2020distilling}
Zizhao Zhang, Han Zhang, Sercan~O Arik, Honglak Lee, and Tomas Pfister. 2020.
\newblock Distilling effective supervision from severe label noise.
\newblock In \emph{Proceedings of the IEEE/CVF Conference on Computer Vision
  and Pattern Recognition}, pages 9294--9303.

\end{thebibliography}
\bibliographystyle{acl_natbib}

\end{document}